# The emerging field of language dynamics

Søren Wichmann

Department of Linguistics, Max Planck Institute for Evolutionary Anthropology

& Faculty of Archaeology, Leiden University

wichmann@eva.mpg.de

Abstract

Large linguistic databases, especially databases having a global coverage such as *The World Atlas of Language Structures* (Haspelmath et al. 2005), The Automated Simility Judgment Program (Brown et al., n.d.) or *Ethnologue* (Gordon 2005) are making it possible to systematically investigate many aspects of how languages change and compete for viability. Agent-based computer simulations supplement such empirical data by analyzing the necessary and sufficient parameters for the current global distributions of languages or linguistic features. By combining empirical datasets with simulations and applying quantitative methods it is now possible to answer fundamental questions such as 'what are the relative rates of change in different parts of languages?', 'why are there a few large language families, many intermediate ones, and even more small ones?', 'do small languages change faster or slower than large ones?' or 'how does the borrowing of words relate to the borrowing of structural features?'

**1. The emerging field of language dynamics**

Throughout most of the history of linguistics, when scholars have taken a broad look at the world's languages the questions they have asked have been of a phylogenetic nature: How are these languages related? What can we say about their origins? In the 1990's some linguists began to ask new types of questions regarding the world's languages as a whole. Nichols (1992) shifted the attention from the historical roots of languages to the historical roots of the structural features that make up languages, noting distributional patterns that range across continents and require explanations in terms of prehistoric interaction and migration; and Nettle (1999a) shifted the attention from finding origins of language families to explaining their current distributions with attention to geographical and socio-economic factors. Looking at languages on a global scale entails the gathering and analysis of large datasets, and quantitative and statistical approaches come into play. Finally, it has now been realized that languages make up ecological systems whose elements show distributional behaviors that may be approximately described by simple mathematical functions (Zanette 2001, Wichmann 2005). The fact that systems emerge from the apparent stochastic behavior of their elements invites computer simulations as a natural additional tool for testing models and hypotheses, and the field of linguistics is therefore now attracting the attention of scholars who are trained in the application of such computational methods, many of them physicists. The interest of this contingent of scholars appears to have been aroused by papers on the dynamics of language extinction by Abrams and Strogatz (2003) and Sutherland (2003) that have appeared in the high-profile journals *Science* and *Nature*, and is part of a larger trend among physicists to extend the application of their methods to social phenomena (Castellano et al. 2007 provide a broad review). Currently every second week or so a paper is published which looks at quantitative aspects of language change or applies computer simulations to investigate how languages interact (cf. Schulze et al. 2008 for a recent review). Thus, barring some predecessors, a new field, which might be designated 'language dynamics', has begun to take shape just over the last 4 years.

In the remainder of this paper I will provide a brief overview of the emerging field of language dynamics, successively focusing on data, methods, and results. Finally a very brief view towards potentially interesting new research areas is provided.

## 2. Data

Much of the research reported here has been made possible through the publication of *The World Atlas of Language Structures*, edited by Haspelmath et al. (henceforth WALS). It contains 57,916 data points from 2560 languages, which are presented in maps showing the distributions of typological language features. Some data points, however, are combinations of others, and others do not relate directly to the rest (i.e. features of writing systems, sign language features, and the paralinguistic use of clicks). Excluding such features not relating directly to spoken languages there are 138 features. While the amount of data is impressive, only a fraction is useful for broader, comparative purposes. For instance, for 1556 languages less than 20 features are attested and only for 230 languages are 60 or more features attested. Thus, only a few hundred languages may be considered well attested. Moreover, errors will naturally creep into a database consisting of data collected by scholars who are not specialists in the languages from which the data are drawn (in other cases apparent conflicts between datapoints turn out to result from different definitions used by different authors, cf. Cysouw [forthc.] for an example). An online version of WALS is expected to appear in the near future, possibly designed such that it may be expanded through the participation of interested contributors (Haspelmath, n.d.). Several other typical databases have been made available online, the largest of which is Jazyki Mira (Languages of the World), which covers close to 400 Eurasian languages and has some 1.2 million data points (Polyakov and Solovyev 2006). Whereas the features of WALS can take anywhere from 2 to 9 values, depending on the way a given author has chosen to encode the information, Jazyki Mira exclusively consists of binarily encoded features, often hierarchically organized (e.g. presence/absence of a certain group of vowels at a higher level and presence/absence of a certain type of vowel at a lower level). This sort of redundancy in part accounts for the enormous amount of data points, but so does the consistency with which as many features as possible are attested for the languages included in the database. The online database still has limited accessibility and is moreover entirely in Russian, but it should become open in the near future, and an English version is in preparation. Examples of online databases limited to specific structural features of languages are the UCLA Phonological Segment Inventory Database[1], Baerman et al. (2002), or Gast et al. (2007). There are concerns among linguists for developing an infrastructure to facilitate the combination of different databases[2], and the first online system for querying several databases simultaneously, The Typological Database System project, has just been launched by a Dutch research group.[3]

For systematic and computationally supported studies of the lexicon across the world's languages a comprehensive set of electronic dictionaries organized according to meanings of lexemes is desirable. While dictionaries are available for thousands of languages no such resource exists, however. The Intercontinental Dictionary Series project founded by Mary Ritchie Key and continued by Bernard Comrie[4] has the desired standardized electronic format and contains up to 1,310 lexical entries per language; the current number of languages represented, however, is only around 250. Another project, the Automated Similarity Judgment Program (ASJP), described in Brown et al. (n.d.),[5] has set out to gather short word lists for the purpose of an automated and consistent classification of the world's languages as well as for statistical investigations of various kinds. Initially 100-item lists, using the 'Swadesh list' (e.g. Swadesh 1971) were collected for 245 languages. On the basis of these, the relative stabilities of the items were determined and a reduced list of 40 items selected. At the time of writing, the project members have added many 40-item lists to the original 245 100-item lists, and the total number of languages processed exceeds 1400. The goal is to make the coverage as comprehensive as at all possible. The ASJP lists have made possible large-scale investigations of language dynamics that could not earlier have been undertaken.

---

[1] Downloadable as a zipped file from from <http://www.linguistics.ucla.edu/faciliti/sales/upsid.zip>.
[2] E.g. E-MELD (http://www.e-meld.org/index.cfm) and GOLD (http://www.linguistics-ontology.org/).
[3] See <http://languagelink.let.uu.nl/tds/index.html>.
[4] See <http://lingweb.eva.mpg.de/ids/>.

For studies involving all the world's languages, a list of these languages, the number of people who speak them as first languages, their locations, and their genealogical classification are necessary, basic pieces of information. The currently best overall catalogue is Gordon et al. (2005), henceforth *Ethnologue*. A drawback of this catalogue is that it is made for the practical purpose of guiding missionary activities. Thus, it excludes most extinct languages and often is not very rigorous with respect to distinctions between what counts as a dialect and what counts as a language or critical with respect to the genealogical classifictions adopted. Efforts are under way for a more comprehensive catalogue which will remedy the deficiencies of *Ethnologue*[6], but so far *Ethnologue* is the best single index to the world's languages.

---

[5] See also <http://email.eva.mpg.de/~wichmann/ASJPHomePage.htm> for updates on the project and links to papers and other materials.
[6] A meeting 28 June, 2007, at the Max Planck Institute for Evolutionary Anthropology, was devoted to discussing plans for a new catalogue of the world's languages, cf. <http://email.eva.mpg.de/~haspelmt/cat.html> for a program and some downloadable contributions.

## 3. Methods

To date, the study of language dynamics has concentrated on how languages change over time and how some languages may go extinct while others thrive. A traditional method for studying how languages change is the comparative method, where early language stages (proto-languages) are reconstructed by comparing related languages and making inferences using knowledge or intuitions about how languages change. This method is often supplemented with a view to geographically contiguous languages that may have contributed to bringing about changes in the languages focused on through diffusion. Often it is difficult to tease apart internal, spontaneous changes from changes that have taken place as a result of outside influence. A solid job of reconstruction requires years of work and is useful for clarifying how the languages studied have come to look the way they look and for enabling the reconstruction of aspects of the culture of proto-speakers. But the method does not lead to broad generalizations about language dynamics because it applies to one language family at a time and is entirely qualitative. In contrast, comparisons across languages on a global scale using the kinds of data described in the previous section, allow for both generalizations and statistical tests of significance.

Empirical investigations may be supplemented by computational modeling of language dynamics. The models used should have a certain degree of realism, but should not try to imitate a complicated reality. Even if linguists sometimes react negatively to this, it is important to operate with a minimum of parameters such that it is possible to clearly identify the contributions of different ingredients of the model to a given result. Discoveries of systematic, quantitative distributions involving the world's languages provide yardsticks for the degree of realism of simulations of global linguistic diversity. For instance, several simulations have attempted to attain the distribution of language family sizes (as measured in the number of languages per family) plotted by Wichmann (2005) and/or the distribution of language sizes, measured in speaker populations, plotted by Sutherland (2003). The hope is that as more and more quantifiable relations in and among languages are discovered and simulation models are developed which can adequately replicate these distributions, the simulation models will of necessity become more and more adequate as models of actual languages, and could therefore be employed for purposes beyond the ones for which they were designed.

Fours classes of models have been applied. One seeks to approximate the development of linguistic diversity through differential equations and does not operate with languages as having internal structure (Abrams and Strogatz 2003, Nettle 1999c). Another studies the interaction among speakers within a simulated space (a lattice) and also does not operate with any internal language structure (de Oliveira et al. 2006a,b, Patriaraca & Leppännen 2004, Pinasco and Romanelli 2006, Tuncay 2007). A third also studies language dynamics in a simulated space and has some simple way of representing language structure, typically as a string of binary features (bitstrings) (Schulze and Stauffer 2005, Kosmidis et al. 2005, Stauffer et al. 2006, de Oliveira n.d., Teşileanu and Meyer-Ortmanns 2006). Finally, a fourth class of models has elaborate structures for simulating languages, but no component for simulating the interaction among languages and issues of global linguistic diversity. Such models, which are numerous in the field of computational linguistics, fall outside the scope of this review since they are not applied to issues of language dynamics understood as including the interaction among languages.

The four classes of model can be summarized in a 2x2 chart as in figure 1.

PLEASE INSERT FIGURE 1 AROUND HERE

The richest and most versatile type of model will operate with both a space of interaction and an internal structure.

A space of interaction may be specified as a geographical space where features such as the effects of geographical distances among languages or physical barriers among them are in the focus of the investigation (Holman et al. 2007, Schulze and Stauffer 2007), or it may be specified more abstractly as a network of interaction such as scale-free networks (Barabási and Albert 1999) or other kinds of networks (e.g. Ke et al., in press), depending on which sort of issue one sets out to investigate. Different sociological models have been applied in different papers, including the different models of Axelrod (1997), Latané (1981), and Nowak et al. (1990). Finally, parameters deriving from basic linguistic knowledge about language dynamics, such as language shift, diffusion, and internal change, have been standard ingredients in much of the work.

This section has briefly sketched how language dynamics have been studied over the past few years. In the following section I shall highlight some of the results that I find most interesting as a linguist.

## 4. Some results

### 4.1. Stability.

It has long been a desideratum to be able to measure how fast different features of language tend to change relative to one another. Some authors who have ventured statements about stabilities of typological features include Nichols (2003) and Croft (1996), and Nichols (1995) suggests different concrete ways of measuring stabilities—what we might call stability metrics. In Wichmann and Holman (n.d.-b) this line of inquiry is broadened to include the entire WALS dataset and different metrics are tested against simulations where there were preset (known) rates of change in languages characterized by a number of features similar in structure and quantity to the number of WALS features. The metric that performed best on this simulated dataset worked in the following way. First we look at related languages in the WALS database feature by feature. We count for all possible pairs of related languages the cases where the given feature has the same value. For each feature we divide the number of related language pairs that have the same value for the given feature by the number of pairs compared. This proportion says something about the degree to which a given feature tends to have a similar value among related languages, which translates into how stable it is. However, it might be the case that a feature value is widely shared among languages because it is simply typical of the world's languages or has been widely diffused. For this reason we also divide the number of unrelated language pairs that share values for a given feature with the number of pairs compared and the resulting figure is now subtracted from the figure obtained for related languages. That gives us a stability measure that also takes into account universality and diffusion (the figure is modified further to balance contributions of language families of different sizes and so on, but these are minor technical details). The results confirmed some of the estimates in the literature, for instance that the subject-verb-object word order is a highly stable (Nichols 2003: 286) or that the presence/absence of definite articles is a highly unstable (Croft 1996: 2006-7), but in a few cases earlier estimates were contradicted, for instance the statement of Nichols (2003: 295) that ergativity is unstable. What explains the stability of some features as opposed to others is presently not clear. It may have to do with how integrated a given structural feature is with other features, i.e. how close it is to the "genius" of a language (Sapir 1970[1921]) or it may have to do with frequencies (Lieberman et al. 2007), but neither hypothesis is easy to investigate.

Another finding of Wichmann and Holman (n.d.-b) is that, barring a few highly unstable features, typological traits on average have a retention rate which is roughly the same as the 0.86 retention rate per 1000 years estimated by Swadesh (1955) for core vocabulary.

This study should, and likely soon will be, replicated on a different set of data, such as Jazyki Mira.

*4.2. Do population structures affect rates of change?*

An early study introducing computer simulations in order to investigate a problem relating to language dynamics was Nettle (1999d). Here is the question is posed whether small languages tend to change faster than large ones, and the question is answered in the affirmative, quite in line with the intuitive feeling that it should be easier for language changes to spread throughout a smaller than through a larger population. In Nettle's simulation model, which is based on Nowak et al (1990), the impact of a linguistic variant is a function of the statuses of the individuals using this variant, their social distance from the learner, and their number. Wichmann et al. (n.d.) recently attempted to test Nettle's conclusions using a different model. Here individuals are connected in a scale-free network (Barabási and Albert 1999), where the impact of a certain individual increases with a probability which is proportional to the impact that the individual already has had. Social distances correspond to distances among individuals in the network. Moreover, differently from Nettle's model of just one language with two competing features, Wichmann et al. operate with many languages having several features. The results are different when one assumes that diffusion only takes place among neighbours in the network (local version) or when it can take place between any nodes (global version). In the local version there is no dependence between the rate of change and the population size, whereas in the global version such a dependence is seen, provided that the rate of diffusion is high enough. Using empirical data from WALS and *Ethnologue* a statistically significant effect supporting Nettle's claim was found, but the effect was much smaller than in his simulations. This study is a good example of how simulations and empirical data can shed mutual light on one another.

*4.3. Lateral and vertical transmission*

General features of language structure are highly prone to diffuse. A clear result from the inspection of WALS maps and statistical investigations of the data that they display is that any feature, if it exists in a given area, may diffuse. Holman et al. (2007) plot the amount of dissimilarities among respectively related and unrelated languages against geographical distances, showing that a similar relationship exists: for both groups dissimilarity increases with distances, but related languages are—not surprisingly—more similar on average than unrelated languages at any given distance. For languages that are around 6000 km removed from one another tend to be maximally dissimilar, and the amount of dissimilarity does not grow beyond this point, suggesting that the range of diffusion roughly lies within 6000 km. Simulations where the rates of diffusion, migration, language shift, and change were varied showed that none of these factors can cause unrelated languages to be more similar on average than related languages. This study averaged over many languages. For individual pairs the situation can be quite different, with certain unrelated languages being extremely similar and certain related languages quite different. Examples of extreme cases, identified in Wichmann and Holman (n.d.-a) are the two Niger-Congo languages Zulu and Ijo, which share only 28.8% similarities in terms of WALS features and the unrelated languages Vietnamese (Austro-Asiatic family) and Thai (Tai-Kadai family), which are 80.9% similar. A closely related pair such as Russian and Polish is less than one percent more similar than Vietnamese and Thai. This means that a language pair has to be as similar as Russian and Polish before one can be certain that they are related. Thus, typological similarity is not a good predictor of relatedness. On the other hand, if languages are very dissimilar one can be pretty certain that they are not generally thought be related. None of the language pairs that are less similar than Zulu and Ijo are related according to generally accepted classifications, and only very few pairs of related languages are less than 40% similar.

The power of diffusion of typological features, then, makes it hard to use such features to establish genealogies. In Dunn et al. (2005: 2072) typological data were used to make the claim that "Papuan languages show an archipelago-based phylogenetic signal (...). The most plausible hypothesis to explain this result is the divergence of the Papuan languages from a common ancestral stock." The authors, however, had only shown that the distribution of similarities among the Papuan languages were consistent with the geographical distances among them. As mentioned above, however, both related and unrelated languages are sensitive to geographical distance in the amount of similarity they exhibit, so the claim that evidence had been found for a phylogenetic relationship was completely unfounded. In fact, in a more recent paper, written in response to a critique by Donohue and Musgrave (2007), the authors now admit that they are "unable to tease ancient contact and phylogeny apart" (Dunn et al. 2007: 401). Wichmann and Saunders (2007) also looked into the use of typological features for making phylogenetic inferences, but took a more cautious approach, seeking proper methodological strategies rather than making spectacular empirical claims. They argued for the necessity of determining which typological features are the more stable ones so as to use only those, and pointed out that the way features are encoded and the choice of tree-generating algorithm was also important. In addition, it was suggested that a combination of typological and lexical features might bring the hope of extending the time depth at which phylogenies may be established. Recently, more results towards the refinement of the suggested methodology have come about. The particular way of measuring stabilities, originally developed in Wichmann and Kamholz (in press), has been superceded by Wichmann and Holman (n.d.), which was summarized in 4.1 above; and Holman et al. (n.d.) have added more substance to the claim that a combination of lexical and typological data may yield better results in terms of accuracy of phylogenies that they can produce than either could alone. Good results are obtained when languages are classified according to the amount of cognates shared on a list of the 40 most stable items on the list of meaning originally developed by Swadesh (1955), but even better results are obtained when typological similarities are also taken into account. A weighting should be produced such that information from the lexicon feeds into about three fourths of each similarity measure and information from typology accounts for one fourth of the measure. Using as many typological features as possible gives the best results, but close to optimal results are obtained using only the 40 most stable typological features.

*4.4. Computational simulations of language competition*

Simulations are most meaningful when supplemented by empirical data, but driven by a specific hypothesis that the empirical data for one reason or the other cannot shed full light on—as in the studies summarized in the previous three subsections. When results come from simulations alone it is harder to assess their validity. Nevertheless, it is possible to draw some generalizations from the work in the area of pure simulation. In a model of the development of global linguistic diversity one can assume several separate 'inventions' of languages or a monolithic model with a single proto-World language. In the monolithic model the degree of ensuing diversity is highly dependent on the rate of change posited. For low rates of changes the original languages and variants thereof will continue to dominate, whereas for higher rates of change a high diversity, similar to what we find in reality, will ensue. When one starts with several different random languages the resulting amount of diversity is similarly dependent upon the rate of change, with more than half of the population eventually speaking just one language for slow rates of change. The amount of time it takes before this dominance of one language sets increase (logarithmically, roughly) with the population size (Schulze and Stauffer 2006). The effects of language shift and diffusion can be blocked by physical barriers, making it possible for languages to remain permanently distinct (Schulze and Stauffer 2007). While most simulation work has looked at agents as monolingual, bilingualism has also been simulated (e.g. Castelló et al. 2006). An interesting result is that the growth of a lingua franca may be speeded up considerably if it assumed that speakers migrate (Schulze et al. 2008). These are some concrete results of simulations which, as said, are somewhat hard to evaluate. But once such simulations are brought into the purview of a concrete research question they may help shed light on the situation. One may begin to ask "what if…" questions.

More central to the computational enterprise than specific research questions like the ones just exemplied has been the development of a model that can capture distributions found in reality and is therefore expected to be efficient when put to the task of clarifying concrete questions. To date, the model which has been most adequate in capturing the present distribution of language sizes and languages family sizes, as measured by the number of languages in each of the world's language families, is a variant of the so-called 'Viviane' model, informally named after the Brazilian physicist Viviane de Oliveira and first presented in de Oliveira et al. (2006a). This model did not operate with languages as having any internal structure—languages were simply represented by a number for each. In de Oliveira et al. (n.d.), following de Oliveira et al. 2007, this model was combined with the so-called 'Schulze' model of Schulze and Stauffer (2005) where language structure is represented as a set of features that can take different values (originally only the values 0 or 1 were allowed for, but later implementations, such as Holman et al. 2007, have allowed for more possible values). In the new model world geography is simulated as a lattice (grid). Initially only the central point is occupied by speakers of just one language. Then agents begin to migrate, and every time a new lattice site is occupied there is a certain probability that the language changes in one of its features and is then defined as a new language. Moreover, there is a certain probability that the language becomes the ancestor of subsequent languages. The different probabilities may be fine-tuned to give just the right distributions of language and family sizes, but, interestingly, the general shapes of the distributions also remain the same as in reality independently of the parameter settings. So the combined 'Viviane-Schulze' model seems to be a suitable one for further investigations of questions of phylogenetic relations among languages and the development of linguistic diversity.

In all the simulations the point is to uncover the statistical properties of language interaction that produce effects independently of whatever contingencies might have occurred in prehistory. For instance, a model can predict that at a certain stage in prehistory there will begin to be just a few relatively large families, many intermediately sized ones, and even more small ones, as in present-day reality (Wichmann 2005, Stauffer et al. 2006). But the reason why it is one particular language which is the most succesful at some point in prehistory will not be the same as the reason for the success of other language later on. For instance, technological advantages that may be involved could be different. It seems difficult to reconcile the statistical approach of physicist who are used to think in terms of, say, random movements of particles with the approach of linguists or archaeologists (e.g. Bellwood and Renfrew 2003 and papers therein), who seek particular determinants, such as the spread of agriculture or other, for the present distribution of languages. Nevertheless, within their respective limitations, both approaches are valid.

**Outlook**

The present review has been quite selective. There are other areas not touched upon here which could be considered  as having to do with language dynamics and where the combination of empirical databases and computational approaches have been or could be employed, for instance language evolution, dialectology or language acquisition. I have also been vague in my characterization of the 'field' of language dynamics. The fact is that it is difficult to characterize and even more difficult to define such a field. But I see this as a sign of health. When, in science, something is happening and we don't quite know what it is, this is usually because what's happening is important and will have a lasting impact.

FIGURE CAPTIONS

Figure 1. Classes of computational models for languages

FIGURES

Figure 1:

| Class 1 | Class 2 |
|---|---|
| -space of interaction | +space of interaction |
| -internal structure | -internal structure |
| Class 3 | Class 4 |
| +space of interaction | -space of interaction |
| +internal structure | +internal structure |